\documentclass{article} 
\usepackage[utf8]{inputenc}
\usepackage{graphicx}
\usepackage{authblk}
\usepackage[export]{adjustbox}
\usepackage{comment}
\usepackage{multirow}
\usepackage{mathtools,xparse}
\usepackage[table,xcdraw]{xcolor}
\usepackage{hyperref}
\usepackage{amssymb}
\usepackage{amsmath}
\usepackage{neuralnetwork}
\usepackage{booktabs}
\usepackage{fancyvrb}
\usepackage{caption}
\usepackage{subcaption}
\usepackage{float}
\usepackage{fullpage}
\usepackage{todonotes}
\usetikzlibrary{positioning}

\newcommand{\com}[1]{{\bfseries \color{red} #1}} 

\newcommand{\withLtwo}[1]{}

\newcommand{\argsort}{\mathtt{argsort}}
\newcommand{\argmax}{\mathtt{argmax}}

\newtheorem{definition}{Definition}

\title{On Neural Network Equivalence Checking using SMT Solvers\thanks{\textbf{Acknowledgment}: This work has received funding from the European Union’s Horizon 2020 research and innovation programme under grant agreement No 956123.}}
\author[1]{Charis Eleftheriadis}
\author[1]{Nikolaos Kekatos}
\author[1]{Panagiotis Katsaros}
\author[2,1]{Stavros Tripakis}

\affil[1]{School of Informatics, Aristotle University of Thessaloniki, Greece}
\affil[2]{Khoury College of Computer Sciences, Northeastern University, Boston, US}

{
    \makeatletter
    \renewcommand\AB@affilsepx{: \protect\Affilfont}
    \makeatother

    \affil[ ]{\rule[12pt]{0pt}{0pt}e-mails:}

    \makeatletter
    \renewcommand\AB@affilsepx{, \protect\Affilfont}
    \makeatother

    \affil[1]{\{celefther,nkekatos,katsaros\}@csd.auth.gr}
    \affil[2]{stavros@northeastern.edu}
}

\date{\today}

\begin{document}

\maketitle

\begin{abstract}

\noindent
Two pretrained neural networks are deemed equivalent if they yield similar outputs for the same inputs. Equivalence checking of neural networks is of great importance, due to its utility in replacing learning-enabled components with equivalent ones, when there is need to fulfill additional requirements or to address security threats, as is the case for example when using knowledge distillation, adversarial training etc. SMT solvers can potentially provide solutions to the problem of neural network equivalence checking that will be sound and complete, but as it is expected any such solution is associated with significant limitations with respect to the size of neural networks to be checked. This work presents a first SMT-based encoding of the equivalence checking problem, explores its utility and limitations and proposes avenues for future research and improvements towards more scalable and practically applicable solutions. We present experimental results that shed light to the aforementioned issues, for diverse types of neural network models (classifiers and regression networks) and equivalence criteria, towards a general and application-independent equivalence checking approach.

\end{abstract}

\section{Introduction}

For two pretrained neural networks of the same or different architectures, the problem of equivalence checking concerns with checking whether the networks yield similar outputs (fit for purpose) for the same inputs. The exact definition of the problem can refer to various ``equivalence'' criteria, depending on the specific neural network application (e.g. classifier, regression, etc.), while the motivation behind the quest for a solution is fundamental, for a series of recent developments in machine learning technology.

More specifically, we refer to knowledge distillation~\cite{Hinton2015DistillingTK}, i.e. the process of transferring knowledge from a large neural network to a smaller one that may be appropriate for deployment on a device with limited computational resources. Another area of interest includes the techniques, widely known under the term regularization~\cite{Kukacka2017RegularizationFD}, which aim to lower the complexity of neural networks, in order to show better performance during inference time, when they process data that are not in the training data set (avoid data overfitting). Moreover, neural network models in systems or programs with learning-enabled components~\cite{Christakis2021} may have to be updated for a number of reasons~\cite{Paulsen_2020}; for example, security concerns such as the need to withstand data perturbations (e.g. adversarial examples), or possibly incomplete coverage of the neural network's input domain.

In all aforementioned cases, we usually require a different neural network model than the original one, which is expected to comply with respect to some given criterion of ``equivalence'', depending on the specific neural network application.

This work presents the experience from our attempt to address the aforementioned problem based on the use of Satisfiability Modulo Theory (SMT) solvers, which provide certain advantages, as well as limitations that justify the need for further research efforts. Among their advantages, we stress their potential to deliver sound and complete verification procedures for the equivalence between two neural networks. Regarding the limitations, we focus on their inability to scale towards solving the problem for real-size neural networks like the ones referred in well-known benchmarks, such as the neural network architectures for the MNIST dataset. We study the scalability bounds of our initial SMT-based encoding for our ``equivalence'' criteria, with respect to the neural network model parameters and the number of derived SMT variables.

More concretely, this article introduces:
\begin{itemize}
    \item the problem definition of equivalence checking based on various criteria that may be appropriate, for different neural network applications 
    \item an approach to reduce the equivalence checking problem to a logical satisfiability problem based on our SMT-based encoding
    \item experimental results including (i) sanity checks of our SMT-based encoding, as well as (ii) equivalence checks for three diverse neural network applications covering the cases of classifiers and regression models
\end{itemize}
Section~\ref{sec:2} lays the background of our work by providing a formal definition of neural network models. In Section~\ref{sec:3}, we propose diverse ``equivalence'' criteria for the wide range of common neural network applications and we formally define the problem of equivalence checking. Section~\ref{sec:4} presents our SMT-based encoding for reducing the problem of equivalence checking to a logical satisfiability problem. Section~\ref{sec:5} includes the experimental results and their interpretation. In Section~\ref{sec:6} we review the related work and the paper concludes with a summary on our contributions and the future research prospects.

\vspace{3mm}

\section{Preliminaries: Neural Networks}
\label{sec:2}

\subsection{Notation}

The set of real numbers is denoted by $\mathbb{R}$.
The set of natural numbers is denoted by $\mathbb{N}$. Given some $x\in \mathbb {R}^n$ and some $i\in \{1,...,n\}$, $x(i)$ denotes the $i$-th element of $x$.

\subsection{Neural Networks}

In general, a neural network (NN) can be defined as a function: 
\begin{equation}
    \label{eq1}
f: I \to O 
\end{equation}
where $I \subseteq \mathbb {R}^n$ is some input domain with $n$ \emph{features} and $O \subseteq \mathbb{R}^m$ is some output domain.

For a neural network image classifier, we typically have $ I = [0, 255]^n \subseteq \mathbb{N}^n$ and a labeling function 
$L: \mathbb{R}^m \to \mathbb{N}$
that maps each $ y \in O$ to some label $l \in \mathbb{N}$. For neural networks solving regression problems, we will have $I \subseteq \mathbb {R}^n$ and no labelling function.

The above definition of neural networks is purely semantic. Concretely, a neural network is structured into layers of \emph{nodes} (neurons), which may include one \emph{hidden layer} ($H$) or more, beyond the layers of input ($I$) and output nodes ($O$). Nodes denote a combination of affine value transformation with a piecewise linear or non-linear \emph{activation function}. Value transformations are \emph{weighted} based on how nodes of different layers are connected with each other, whereas an extra term called \emph{bias} is added per node. Weights ($W$) and biases ($b$) for all nodes of a neural network are the network's {\em parameters} and their values are determined via {\em training}. 

Since every layer is multidimensional we use vectors and/or matrices to represent all involved operations. Let $\mathbf{x} \in \mathbb {R}^{1 \times n}$ be the matrix denoting some $x \in I$. For a hidden layer with $r$ nodes, $\mathbf{H}^{1 \times r}$ represents the output of the hidden layer. Assuming that the hidden and output layers are fully connected, we denote with $\mathbf{W}^{{(1)}} \in \mathbb {R}^{n \times r}$ the hidden layer weights and with $\mathbf{b}^{{(1)}} \in \mathbb {R}^{1 \times r}$ the biases associated with its nodes. Similarly, the output layer weights are denoted by $\mathbf{W}^{{(2)}} \in \mathbb {R}^{r \times m}$, where $m$ refers to the number of output layer nodes, and $\mathbf{b}^{{(2)}} \in \mathbb {R}^{1 \times m}$ denotes the corresponding biases. Then, the output $y = f(x)$ of the neural network is given by $\mathbf{y} \in \mathbb {R}^m$ where $\mathbf{y}$ is computed as:
\begin{align}
&\mathbf{H} = \alpha(\mathbf{x} \, \mathbf{W}^{{(1)}} + \mathbf{b}^{{(1)}})&& \label{eq:2} \\
&\mathbf{y} = \alpha^\prime(\mathbf{H} \, \mathbf{W}^{{(2)}} + \mathbf{b}^{{(2)}}) \label{eq:3} &&
\end{align}
where $\alpha(\cdot),\alpha^\prime(\cdot)$ denote the {\em activation functions} (e.g. $sigmoid$, $tanh$, $ReLU$ etc.) that are applied to the vectors of the hidden and the output layers element-wise. For example,  a commonly used activation function is the ReLU, which is defined, for $\chi \in \mathbb{R}$, as:
\begin{align}
    ReLU(\chi)= 
\begin{cases}
    \chi,  & \text{if } \chi\geq 0\\
    0,  & \text{otherwise}
\end{cases}
\end{align}

Another example is the {\em hard $tanh$} function~\cite{collobert:2004b} that serves as the output layer activation function of the neural networks that we have trained for a regression problem that we addessed in the experiments of Section~\ref{sec:5}. For $\chi \in \mathbb{R}$, hard $tanh$ is defined as:

\begin{align}
    HardTanh(\chi)= 
\begin{cases}
    1,  & \text{if } \chi > 1 \\
    -1,  & \text{if } \chi < -1 \\
    \chi,  & \text{otherwise}
\end{cases}
\end{align}


\paragraph{Generalization to multiple hidden layers}

The  neural network definition provided above can be easily generalised to any number of hidden layers $H, H^\prime, H^{\prime\prime}, \cdots$. We consider that weights and biases for all layers and nodes are fixed, since we focus on equivalence checking of neural networks {\em after} training.

\paragraph{Example}

Consider a simple feedforward NN  with two inputs, two outputs and one hidden layer with two nodes. The selection of the weights and biases is done randomly, the activation function of the hidden layer is ReLU while there is no activation function for the output layer.
A drawing of the NN is shown in Figure~\ref{fig:NN11}. For this example, equation (\ref{eq:2}) takes the form: 
\begin{align*}
\begin{bmatrix}
x_1 & x_2
\end{bmatrix}
\cdot
\begin{bmatrix}
W_{11} & W_{12} \\
W_{21} & W_{22}
\end{bmatrix}
+
\begin{bmatrix}
b^{(1)}_{1} & b^{(1)}_{2} 
\end{bmatrix}
&=
\begin{bmatrix}
x_1 & x_2
\end{bmatrix}
\cdot
\begin{bmatrix}
-2 & 1 \\
1 & 2
\end{bmatrix}
+
\begin{bmatrix}
1 & 1
\end{bmatrix}
\\
&=
\begin{bmatrix}
-2 \cdot x_1 + x_2  + 1  & x_1 + 2 \cdot x_2 + 1 
\end{bmatrix}
\end{align*}

Denote the result of the affine transformation of the neural network's hidden layer by:
\begin{align*}
\begin{bmatrix}
z_1  & z_2
\end{bmatrix}
&=
\begin{bmatrix}
-2 \cdot x_1 + x_2  + 1  & x_1 + 2 \cdot x_2 + 1 
\end{bmatrix}
\end{align*}

Then, we  get:
\begin{align*}
\mathbf{H} &=
\begin{bmatrix}
h_1  & h_2
\end{bmatrix}
=
\begin{bmatrix}
ReLU(z_1)  & ReLU(z_2)
\end{bmatrix}
\end{align*}
The output $\mathbf{y}$ of the neural network from equation (\ref{eq:3}) is:
\begin{align*}
\mathbf{y}
=
\begin{bmatrix}
h_1 & h_2
\end{bmatrix}
\cdot
\begin{bmatrix}
2 & -1 \\
-1 & -2
\end{bmatrix}
+
\begin{bmatrix}
b^{(2)}_{1} & b^{(2)}_{2}
\end{bmatrix}
=
\begin{bmatrix}
2 \cdot h_1 - h_2  + 2  & - h_1 - 2 \cdot h_2 + 2 
\end{bmatrix}
\end{align*}

\newcommand{\logiclabel}[1]{\,{$\scriptstyle#1$}\,}
		\newcommand{\linklabelsU}[4]{\logiclabel{+1}}
				\newcommand{\linkRelu}[4]{\logiclabel{ReLU}}
				\newcommand{\linkLinear}[4]{\logiclabel{Linear}}
				\newcommand{\linklabelsB}[4]{\logiclabel{-1}}
				\newcommand{\linklabelsC}[4]{\logiclabel{2}}

		\newcommand{\linklabelsA}[4]{\ifnum0=#2 \logiclabel{+3} \else \logiclabel{-2} \fi}

\newcommand{\inputnum}{2} 
 
\newcommand{\hiddennum}{2}  

\newcommand{\outputnum}{2} 
 
 \begin{figure}[ht!]
\begin{center}
\begin{tikzpicture}
 

    \node[circle,
 label={[align=left]\normalsize \bf Input\\ \normalsize \textbf{Layer}\\~}, 
        text width=4mm,
        text height=3mm,
        fill=orange!30] (Input-1) at (0,-1*1.7) {$x_1^{\textcolor{orange!30}{[1]}}$};

    \node[circle, 
        text width=4mm,text height=3mm,
        fill=orange!30] (Input-2) at (0,-2*1.7) {$x_2^{\textcolor{orange!30}{[1]}}$};


    \node[circle, 
        label={[align=left]\normalsize \bf Hidden\\ \normalsize \textbf{Layer}\\~}, 
        text width=4mm,text height=3mm,
        fill=blue!50,
        yshift=(\hiddennum-\inputnum)*5 mm
    ] (Hidden-1) at (2,-1*1.7) {$z_1^{[1]}$};
 \node[circle, 
        text width=4mm,text height=3mm,
        fill=blue!50,
        yshift=(\hiddennum-\inputnum)*5 mm
    ] (Hidden-2) at (2,-2*1.7) {$z_2^{[1]}$};
 
 \foreach \i in {1,...,\hiddennum}
{
    \node[circle, 
        text width=4mm,text height=3mm,
        fill=blue!50,
        yshift=(\hiddennum-\inputnum)*5 mm
    ] (Hidden_Relu-\i) at (4,-\i*1.7) {$h_{\i}^{\textcolor{blue!50}{[1]}}$};
}

\node[circle, 
    label={[align=left]\normalsize \bf Output\\ \normalsize \textbf{Layer}\\~}, 
    text width=4mm,text height=3mm,
    fill=purple!50,
    yshift=(\outputnum-\inputnum)*5 mm] (Output-1) at (6,-1*1.7) 
    {$y_1^{[2]}$};
    
    \node[circle, 
        text width=4mm,text height=3mm,
        fill=purple!50,
        yshift=(\outputnum-\inputnum)*5 mm
    ] (Output-2) at (6,-2*1.7){$y_2^{[2]}$};

 \draw[->, shorten >=1pt] (Input-1) -- node[above] {$-2$}(Hidden-1);
 \draw[->, shorten >=1pt] (Input-2) --  node[below]{$2$}(Hidden-2);
 \draw[->, shorten >=1pt] (Input-1) --  node[above]{$1$}(Hidden-2);
  \draw[->, shorten >=1pt] (Input-2) --  node[below]{$1$}(Hidden-1);

\foreach \i in {1,...,\hiddennum}
{
    {
        \draw[->, shorten >=1pt]  (Hidden-\i) -- node[above,sloped] {\em \small ReLU}(Hidden_Relu-\i);
    }
}

\draw[->, shorten >=1pt] (Hidden_Relu-1) -- node[above]{2}(Output-1);
\draw[->, shorten >=1pt] (Hidden_Relu-2) -- node[below]{-2}(Output-2);
\draw[->, shorten >=1pt] (Hidden_Relu-2) -- node[above]{-1}(Output-1);
\draw[->, shorten >=1pt] (Hidden_Relu-1) -- node[below]{-1}(Output-2);
\foreach \i in {1,...,\inputnum}
{            
    \draw[<-, shorten <=1pt] (Input-\i) -- ++(-1,0)
        node[left]{};
}
\foreach \i in {1,...,\outputnum}
{            
    \draw[->, shorten <=1pt] (Output-\i) -- ++(1,0)
        node[right]{};
}
 
\end{tikzpicture}

\end{center}
\caption{Feedforward neural network with an input layer of 2 inputs ($x_1$, $x_2$), an output layer with 2 outputs ($y_1$, $y_2$) and no activation function, and 1 hidden layer with 2 neurons and a ReLU activation function. 
	The  values on the transitions correspond to the \emph{weights} and  the upperscript values to the biases.
    \label{fig:NN11}}
\end{figure}
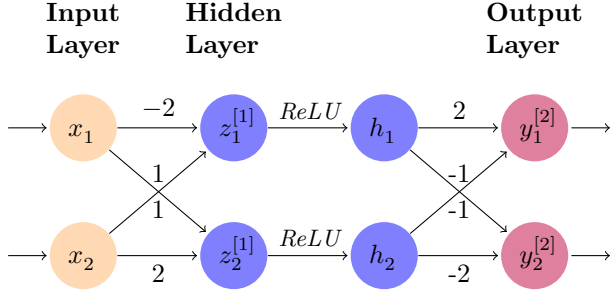 



\section{Strict and Approximate Equivalences for Neural Networks}
\label{sec:3}

In this section, we propose different equivalence relations and we formulate the equivalence checking problem.

\subsection{Strict Neural Network Equivalence}

Strict NN equivalence is essentially functional equivalence:

\begin{definition}[Strict NN Equivalence]
\label{def_strict}
Consider two neural networks $f : I \to O$ and $f' : I \to O$.
We say that $f$ and $f'$ are \emph{strictly equivalent}, denoted $f \equiv f'$, if and only if the following holds:
\begin{equation}
    \label{eq2}
\forall x \in I, f(x) = f'(x)
\end{equation}
\end{definition}

Strict NN equivalence is a true equivalence relation, i.e., it is reflexive ($f \equiv f$ for any NN $f$),
symmetric ($f \equiv f'$ iff $f' \equiv f$), and transitive ($f \equiv f'$ and $f' \equiv f''$ implies $f \equiv f''$).

However, strict NN equivalence can often be a too strong requirement. For example, if we have two classifiers we may want to consider them equivalent if they always select the same top output class, even though they may not order the remaining output classes in the same way. This motivates us to consider the following {\em approximate} notions of equivalence. We remark that these approximate ``equivalences'' need not be true equivalences, i.e., they may not satisfy the transitivity property (although they are always reflexive and symmetric).

\subsection{Approximate Neural Network Equivalences based on Lp Norms}

As usual, we assume that $O \subseteq \mathbb{R}^m$.
Let $\lVert y \rVert_p=norm_p(y)$ denoting the $Lp$-norm of vector $y \in O$, for $norm_p:O \to \mathbb{R}$ with $p=1,2,\infty$. For any two vectors $y, y'\in O$, if $p=1$ we have the widely used Manhattan norm, $L_1(y,y^\prime)=\lVert y - y' \rVert_{1}= \sum_{i=1}^{m} \lvert y(i) - y'(i) \rvert$, which measures the sum of (differences) between two vectors. For $p=2$, we refer to the so-called Euclidean distance $L_2(y,y^\prime)=\lVert y - y' \rVert_{2}= (\sum_{i=1}^{m} \lvert y(i) - y'(i) \rvert^2\large)^\frac{1}{2}$.
Finally, for $p=\infty$ 
we have the $L_\infty$ distance, which measures the maximum change to any coordinate, i.e. it is given as:
\begin{equation}
    \label{eq4}
    L_\infty(y,y^\prime)=\lVert \hspace{1mm} y - y' \hspace{1mm} \rVert_{\infty} = \max (\lvert y(1) - y'(1) \rvert, \dots, \lvert y(m) - y'(m) \rvert )
\end{equation}

Then, we define the following notion of approximate equivalence:

\begin{definition}[$(p,\epsilon)$-approximate equivalence]
\label{def_epsilonapprox}
Consider two neural networks $f : I \to O$ and $f' : I \to O$, $norm_p: O \to \mathbb{R}$, and some $\epsilon > 0$.
We say that $f$ and $f'$ are $(p,\epsilon)$-\emph{approximately equivalent}, denoted $f \sim_{p,\epsilon} f'$, if and only if the following holds:
\begin{equation}
    \label{eq3}
\forall x \in I, \hspace{2mm} \lVert f(x) - f'(x) \rVert_{p} < \epsilon
\end{equation}
\end{definition}
It can be seen that the relation $\sim_{p,\epsilon}$ is reflexive and symmetric.

\subsection{Approximate Neural Network Equivalences based on Order of Outputs}

NN classifiers work essentially by computing output values and then mapping these outputs to specific classes.
For such networks, we may want to consider them equivalent if they always produce the same order of outputs, even though the output values 
might not be the same.
For example, consider two classifiers $f$ and $f'$ over three possible output classes.
Suppose that, for a given input, $f$ produces $(0.3,0.5,0.2)$ whereas $f'$ produces $(0.25,0.6,0.15)$.
We may then consider that for this input the outputs of $f$ and $f'$ are equivalent, since they have the same order, namely,
$2,1,3$ (assuming vector indices start at $1$).
If this happens for all inputs, we may want to consider $f$ and $f'$ (approximately) equivalent. 

To capture the above notion of approximate equivalence, we introduce the $\texttt{argsort}_m$ function, where $m\in\mathbb{N}$:   
\begin{align*}
\argsort_m: \mathbb{R}^m \to \mathcal{Z}_m  
\end{align*}
where $\mathcal{Z}_m \subseteq \{1, 2, 3, \mathellipsis, m \}^m$ is the set of permutations of indices of the $m$ elements.
Then, for a given $s \in \mathbb{R}^m$, $\texttt{argsort}_m(s)$ returns the permutation that sorts $s$ in decreasing order. For example, $\argsort_3(0.3,0.5,0.2) = \argsort_3(0.25,0.6,0.15) = (2,1,3)$.
In the case where two vector values are equal, $\argsort$ orders them from lower to higher index. This ensures determinism of the $\argsort$ function. For example, $\argsort_3(0.3,0.4,0.3) = (1,3,2)$.

\begin{definition}[Top-$k$ $\argsort$ equivalence]
\label{def_argsort}
Suppose $O \subseteq \mathbb{R}^m$.
Consider two neural networks $f : I \to O$ and $f' : I \to O$, and some $k\in\{1,...,m\}$.
We say that $f$ and $f'$ are {\em top-$k$ $\argsort$ equivalent},
denoted $f \approx_k f'$, if and only if
\begin{equation}
    \label{eq4}
\forall x \in I, \forall i\in\{1,...,k\}, \Big(\argsort_m\big(f(x)\big)\Big)(i) = \Big(\argsort_m\big(f'(x)\big)\Big)(i)
\end{equation}
\end{definition}
Top-$k$ $\argsort$ equivalence requires the first $k$ indices of the \texttt{argsort} of the outputs of $f$ and $f'$ to be  equal.  
Top-$k$ $\argsort$ equivalence is reflexive, symmetric, and transitive, i.e., it is a true equivalence relation.

A special case of top-$k$ $\argsort$ equivalence is when $k=1$. 
We call this special case the {\em $\argmax$ equivalence}, with reference to the
 $\argmax$ function which returns the index of the maximum value of a given vector.
For example, $\argmax(0.3,0.5,0.2) = \argmax(0.25,0.6,0.15) = 2$.

\begin{definition}[$\argmax$ equivalence]
\label{def_argmax}
Consider the same setting as in Definition~\ref{def_argsort}.
We say that $f$ and $f'$ are {\em $\argmax$ equivalent}
iff $f \approx_1 f'$.
\end{definition}


\subsection{The Neural Network Equivalence Checking Problem}
\label{sec:eq-problem}


\begin{definition}[NN equivalence checking problem]
Given two (trained) neural networks $f$ and $f'$, and given a certain NN equivalence relation $\simeq\;\in\{\equiv,\sim_{p,\epsilon},\approx_k\}$, and parameters $p,\epsilon,k$ as required,
the {\em neural network equivalence checking problem} (NNECP) is to check whether $f \simeq f'$.
%
\end{definition}

\section{Neural Network Equivalence Checking using SMT Solvers}
\label{sec:4}
Our approach to solving the NNECP is to reduce it to a logical {\em satisfiability} problem. The basic idea is the following. Suppose we want to check whether $f \simeq f'$, for two NNs $f:I\to O$ and $f': I\to O$ and a given NN equivalence relation $\simeq$. We proceed as follows:
(1) encode $f$ into an SMT formula $\phi$;
(2) encode $f'$ into an SMT formula $\phi'$;
(3) encode the equivalence relation $f \simeq f'$ into an SMT formula $\Phi$ such that $f \simeq f'$ iff $\Phi$ is unsatisfiable;
(4) check, using an SMT solver, whether $\Phi$ is satisfiable: if not, then $f \simeq f'$; if $\Phi$ is satisfiable, then $f$ and $f'$ are not equivalent, and the SMT solver also typically provides a {\em counterexample}, i.e., an input which violates the equivalence of $f$ and $f'$.

This idea is based on the fact that the negation of $f \simeq f'$ can be encoded as a formula which asserts that there exist an input $x\in I$ and two outputs $y,y'\in O$, such that $y=f(x)$, $y'=f'(x)$, and $y$ and $y'$ do not satisfy the equivalence conditions imposed by $\simeq$. For example, for the case of strict NN equivalence, checking whether $f\equiv f'$ amounts to checking:
$$
\neg \Big( \exists x \in I, y \in O, y' \in O, \; \; y = f(x) \land y'=f'(x) \land y \neq y' \Big)
$$
This in turn amounts to checking that the formula $y = f(x) \land y'=f'(x) \land y \neq y'$ is unsatisfiable.
In this case, we have $\phi := y = f(x)$, $\phi' := y' = f'(x)$, and $\Phi := \phi \land \phi' \land y \neq y'$.

We proceed to provide the details of building $\phi$ and $\phi'$ for given neural networks, as well as $\Phi$ for the NN equivalence relations defined earlier.


\subsection{Encoding Neural Networks as SMT Formulas}
\label{sec_encodeNN}

\subsubsection{Input variables}

From (\ref{eq:2}), the input of a neural network $f$ is a vector $\mathbf{x} = [ x_1, ..., x_n ] \in \mathbb {R}^{1 \times n}$.
The SMT formula $\phi$ encoding $f$ will have $n$ {\em input variables}, which we will denote $x_1,...,x_n$.

\subsubsection{Encoding input constraints}

Sometimes the inputs are constrained to belong in a certain region. For example, we might assume that the input lies between given lower and upper bounds. In such cases, we can add input constraints as follows:

\begin{align} 
\bigwedge_{j=1}^n  l_j \leq x_j \leq u_j
\end{align}
with $l_j, u_j \in \mathbb{R}$ denoting the lower and upper bounds for the domain of input feature $x_j$. 

\subsubsection{Internal variables}
For each hidden layer, we associate a set of {\em internal variables} $z_i$ corresponding to the affine transformation, and a set of internal variables $h_i$ corresponding to the activation function.

\subsubsection{Constraints encoding the affine transformations}

Consider a single hidden layer of $f$ with $r$ nodes.
Then, from the affine transformation of (\ref{eq:2}), we derive the constraints:
\begin{align} 
\bigwedge_{j=1}^r \left( z_j = \sum_{k=1}^{n} x_k W^{(1)}_{kj} + b^{(1)}_j \right) \label{eq:9}
\end{align}

\subsubsection{Constraints encoding the ReLU activation function}

If the activation function is $ReLU$, then its effect is encoded with the following constraints:

\begin{align}
\bigwedge_{j=1}^r ( z_j \geq 0 \land h_j = z_j) \lor (z_j < 0 \land h_j=0) \label{eq:10}    
\end{align}

\subsubsection{Constraints encoding the hard $tanh$ activation function}

If the activation function is the hard $tanh$, then its effect is encoded with the following constraints:

\begin{align}
\bigwedge_{j=1}^r \large( z_j \geq 1 \land h_j = 1) \lor (z_j \leq -1 \land h_j=-1\large) \lor (-1 < z_j < 1 \land z_j=h_j) \label{eq:10}    
\end{align}

\subsubsection{Other activation functions}

The constraints described so far include atoms of the linear real arithmetic theory~\cite{Kroe08} that most SMT-solvers can check for satisfiability through decision procedures of various degrees of efficiency, for the problem at hand.

Other activation functions than the ones described above are often problematic as there are
limitations regarding the expressions and relations that SMT solvers can handle. For example, most SAT/SMT solvers cannot handle formulas with exponential terms, which is the case of  activation functions such as (not hard) Tanh, Sigmoid, and Softmax. This fact raises the need to find alternative ways to represent such functions, through e.g.  using encodings for ``hard'' versions of these functions found in related works~\cite{PGL-051},~\cite{collobert:2004b}. 
This is the reason we opted for the hard $tanh$ encoding described above.

\subsubsection{Generalization to multiple hidden layers and output layer}

The constraints of equations (\ref{eq:9} - \ref{eq:10}) are generalized to any number of hidden layers, say $H, H^\prime, H^{\prime\prime}, \mathellipsis$ (that consist respectively of $r, r^\prime, r^{\prime\prime}, \mathellipsis$ nodes). 


The neural network encoding is completed with the constraints for the output layer that are derived, for $\mathbf{y} \in \mathbb {R}^m$ from equation (\ref{eq:3}), as previously.

\paragraph{Example (continued)} 


This example contains a single NN and we herein show how the SMT constraints are derived and encoded in case $0\leq x_1 \leq1$ and $0\leq x_2 \leq 1$. For the transformation, we introduce the variables $z_1$ and $z_2$ with  $z_1=-2\cdot x_1+ x_2+1$ and $z_2=x_1+2\cdot x_2+1$. Moving to the activation functions, we have $h_1=ReLU(z_1)=\max(0,z_1)$. We add the constraint  $\{(z_1\geq 0  \wedge h_1=z_1) \vee (z_1<0 \wedge h_1=0)\}$. Similarly, $h_2=ReLU(z_2)=\max(0,z_2)$. We add  $\{(z_2\geq 0  \wedge h_2=z_2) \vee (z_2<0 \wedge h_2=0)\}$. Finally, we add the output constraint $y_1=2\cdot h_1-h_2+2$ and $y_2=-h_1-2\cdot h_2+2$. 

The resulting SMT formula consists of the following constraints 
\begin{align*}
   \phi:=\{&0\leq x_1 \wedge x_1 \leq1 \wedge 0\leq x_2 \wedge x_2\leq 1\wedge z_1=-2\cdot x_1+ x_2+1  
   \wedge ((z_1\geq 0  \wedge h_1=z_1) \vee (z_1<0 \wedge h_1=0)) \nonumber\\ &
   \wedge z_2=x_1+2\cdot x_2+1   \wedge ((z_2\geq 0  \wedge h_2=z_2) \vee (z_2<0 \wedge h_2=0)) \wedge y_1'=2\cdot h_1-h_2+2  \nonumber\\  & \wedge y_2=-h_1-2\cdot h_2+2  \}
\end{align*}

\subsection{ Encoding of the equivalence relation}

In Subsection~\ref{sec_encodeNN} we described how to encode a given neural network $f$ as a formula $\phi$. As mentioned at the beginning of this section, in order to check equivalence of two given neural networks $f$ and $f'$, we need to generate, first, their encodings $\phi$ and $\phi'$ as described in Subsection~\ref{sec_encodeNN}, and second, the encoding of the (negation of the) equivalence relation. The latter encoding is described next.

We assume that the two neural networks $f$ and $f'$ to be compared have the same number of outputs $m$, and we let $\mathbf{y}=(y_1,...,y_m)$ and $\mathbf{y'}=(y_1',...,y_m')$ denote their respective output variables.

\subsubsection{Strict equivalence checking} 
Strict equivalence (c.f., Definition~\ref{def_strict}) requires that $\mathbf{y} = \mathbf{y}^\prime$. As explained in the beginning of this section, the reduction to a satisfiability problem means that we must encode the negation of the above constraint, namely:

\begin{align}
\bigvee_{i=1}^m y_i \ne y'_i
\end{align}

\subsubsection{$(p,\epsilon)$-approximate equivalence checking}

$(p,\epsilon)$-approximate  equivalence (c.f., Definition~\ref{def_epsilonapprox}) requires that $\mathbf{ \lVert y - y^\prime \rVert_{p} < \epsilon}$. Again, we encode the negation:

\begin{itemize}
\item for $p=1$,
\begin{align}
\sum_{i=1}^{m} \lvert y_i - y^\prime_i \rvert \geq \epsilon
\end{align}
\item for $p=2$,
\begin{align}
    (\sum_{i=1}^{m} \lvert y_i - y^\prime_i \rvert^2\large)^\frac{1}{2} \geq \epsilon
\end{align}
\item whereas for $p=\infty$,
\begin{align}
    \bigvee_{i=1}^m \lvert y_i - y^\prime_i \rvert \geq \epsilon
\end{align}
\end{itemize}
The $L_2$ norm is currently not supported in our implementation.

\subsubsection{argmax equivalence checking}
\newcommand{\argmaxis}{{\mathsf{argmaxis}}}

argmax  equivalence (c.f., Definition~\ref{def_argmax}) requires that 
$\textbf{argmax(y)} = \textbf{argmax}\mathbf{(y^\prime)}$.
Again, we wish to encode the negation, i.e., 
$\textbf{argmax(y)} \ne \textbf{argmax}\mathbf{(y^\prime)}$.
This can be done by introducing the macro $\argmaxis(y,i,m)$ which represents the constraint $\argmax(y)=i$, assuming the vector $y$ has length $m$.
Then, $\textbf{argmax(y)} \ne \textbf{argmax}\mathbf{(y^\prime)}$
can be encoded by adding the constraints below:

\begin{align}
\mathop{\bigvee_{i,i'\in \{1,..., m\}}}_{i\ne i'} \argmaxis(y,i,m) \land \argmaxis(y',i',m)
\end{align}

where $\argmaxis$ is defined as follows:
\begin{align}
\argmaxis(y,i,m) := 
\big( \bigwedge_{j=1}^{i-1} y_i > y_j \big)
\land 
\big( \bigwedge_{j=i+1}^m y_i \ge y_j \big)
\end{align}

For example, for $m=2$, we have:
\begin{align*}
\argmaxis(y,1,2) = y_1 \ge y_2 \\
\argmaxis(y,2,2) = y_2 > y_1 \\
\argmaxis(y',1,2) = y'_1 \ge y'_2 \\
\argmaxis(y',2,2) = y'_2 > y'_1 
\end{align*}
and the overall constraint encoding 
$\textbf{argmax(y)} \ne \textbf{argmax}\mathbf{(y^\prime)}$
becomes:
\begin{align*}
(y_1 \ge y_2 \land y'_2 > y'_1)
\lor
(y_2 > y_1 \land y'_1 \ge y'_2)
\end{align*}






\paragraph{Example (continued)} 
Let's assume that there is an additional neural network $f'$ with the same number of inputs and outputs as $f$ that can be encoded via constraints $\phi'$. Below, we show how the complete SMT formula $\Phi$ would be encoded for different equivalence relations:

\begin{align*}
    \Phi_{strict}:&=\{\phi \wedge \phi'\wedge \bigvee_{i=1}^m y_i \ne y'_i \}\\
    \Phi_{(1,\epsilon)-\text{approx}}:&=\{\phi \wedge \phi' \wedge \sum_{i=1}^{m} \lvert y_i - y^\prime_i \rvert \geq \epsilon \}\\
    \Phi_{(\infty,\epsilon)-\text{approx}}:&=\{\phi \wedge \phi' \wedge \bigvee_{i=1}^m \lvert y_i - y^\prime_i \rvert \geq \epsilon \}\\
    \Phi_{argmax}:&=\{\phi \wedge \phi' \wedge \mathop{\bigvee_{i,i'\in \{1,..., m\}}}_{i\ne i'} \argmaxis(y,i,m) \land \argmaxis(y',i',m) \}
\end{align*}

\section{Experimental results}
\label{sec:5}

In this section, we report on a set of experiments on verifying equivalence relations between two NNs. In our experiments we have used the  SMT solver Z3~\footnote{https://z3prover.github.io/api/html/} to check satisfiability of all formulas used to encode NN equivalence. All experiments were conducted on a laptop with a 4-core $2.8$GHz processor and $12$ GB RAM. We experiment with different sizes of NNs, while we conduct equivalence checking for the various equivalence relations defined in Section~\ref{sec:3}. We focus on the two main categories of supervised learning problems, i) classification, and ii) regression. We examine two case studies for classification and one for regression.\par

\vspace{3mm}
\noindent
\textit{Bit-Vec case study -- Classification}\par
\vspace{1mm}
\noindent A bit vector (Bit-Vec) is a mapping from an integer domain to values in the set {0, 1}. For this case study, we consider that the inputs are 10-bit vectors and the targets (labels) are binary: either True (1) or False (0). The models we check for equivalence are Feed-Forward Multi-Layer Perceptrons and we make use of two different architectures. In the first, there is a single hidden layer, while in the second one there are two hidden layers; the networks have the same number of nodes per layer. We experiment with $8$ different models per architecture and for each model we incrementally increase the number of nodes per layer. We train neural networks with the objective to approximate (``learn'') that for 3 or more consecutive 1s in the vector the output label is True, otherwise the output label is False.

\vspace{3mm}
\noindent
\textit{MNIST case study -- Classification}\par
\vspace{1mm}
\noindent The second case study uses the \emph{MNIST} dataset, a popular dataset on image classification. The dataset contains 70,000 grayscale images, from which 60,000 are used for training and the rest 10,000 for testing the models' performance. Every image's size is 28x28 (pixels), while every pixel value is a real number in the range $[0, 1]$. We experiment with the same two architectures used in the Bit-Vec case study and $5$ models per architecture.\par

\vspace{3mm}
\noindent
\textit{Automotive Control -- Regression}\par
\vspace{1mm}
\noindent For the regression case study, the goal is to use neural networks to approximate the behaviour of a Model Predictive Controller (MPC) that has been designed for an automotive lane keeping assist system. The dataset contains 10,000 instances with six features representing different system characteristics obtained by the sensors and the resulting steering angle that the automotive car should follow (target). In Table~\ref{tab:input_features} the features' and target's details and valid value ranges are presented.\par

\begin{table*}[h]\centering
\vspace{0.2em}
\begin{tabular}{@{}lcl@{}}
\toprule
\textbf{Type/Parameter} & \textbf{Answer/Value} & \textbf{Remarks}\\
\midrule
    output/target & $[-1.04, 1.04]$ & steering angle $[-60,60]$ \\
    input range $x_1$ & [-2,2] & $v_x$ (m/s) \\
    input range $x_2$ & [-1.04, 1.04]& rad/s\\
    input range $x_3$ & [-1,1] &m \\
    input range $x_4$ & [-0.8, 0.8]& rad\\
    input range $x_5$ & [-1.04, 1.04] & $u_0$ (steering angle)\\
    input range $x_6$ & [-0.01,0.01] & $\rho$\\
     \bottomrule
\end{tabular}
\caption{Regression Problem Input Characteristics -- Constraints }
\label{tab:input_features}
\end{table*}%

\subsection{Sanity checks}


In this first set of experiments, we have two primary goals to achieve. First, we want to sanity-check our prototype implementation, in order to ensure that it does not contain any bugs and that it is able to provide concrete solutions. Second, we want to perform an empirical scalability study focusing on the  computational demands required for NNs of increasing complexity to be equivalence-checked according to the relations described in Section~\ref{sec:3}.\par

More specifically, we verify two identical neural networks under the various equivalence relations. We apply these checks in the BitVec and MNIST case studies and for two neural network architectures for each of them.  Tables~\ref{tab:Table 2} and~\ref{tab:Table 3} summarize the results for the BitVec case study, while  Tables~\ref{tab:Table 4} and~\ref{tab:Table 5} summarize the results for the MNIST case study. 
In each table, the first column shows the number of nodes per hidden layer per model, the second column the number of trainable parameters per model and the third the total number of variables (``unknowns'') in the formula given to the SMT solver. 
The tables then report the time in seconds (s) that the SMT solver took to check each equivalence relation for each pair of identical neural networks. In the BitVec tables this is the average over 10 runs, with the worst and best case not presented since the standard deviation between observations is below $ 3\%$ in all occasions. In the MNIST tables, each sanity check was conducted once as the neural networks have much bigger size and a longer time is needed for every check to be completed. 

\vspace{3mm}

\begin{table}[h]
\begin{adjustbox}{center}
\begin{tabular}{ccccccc}
\toprule
\multicolumn{1}{p{2cm}}{\centering \textbf{\#nodes\\ per  layer}} &
\multicolumn{1}{p{1.4cm}}{\centering \textbf{\# params}} &
\multicolumn{1}{p{1.4cm}}{\centering \textbf{\# SMT} \\ \textbf{variables}} &   \multicolumn{1}{p{1.4cm}}{\centering \textbf{Strict} \\ \textbf{Equiv.}} & 
\multicolumn{1}{p{1.4cm}}{\centering \textbf{$L_{1}$ }\\ 
\textbf{Equiv.}} 
& \multicolumn{1}{p{1.4cm}}{\centering \textbf{$L_{\infty}$} \\ 
\textbf{Equiv.}} 
& \multicolumn{1}{p{1.4cm}}{\centering \textbf{Argmax} \\ \textbf{Equiv.}} \\ \hline
10 & 132 & 498 & 0.06 & 0.07 & 0.07 & 0.06  \\ \hline 
20 & 262 & 978 & 0.1 & 0.1  & 0.1 & 0.1 \\ \hline
35 & 457 & 1698 & 0.17 & 0.17  & 0.17 & 0.17  \\ \hline
50 & 652 & 2418 & 0.23 & 0.24 & 0.24 & 0.23  \\ \hline
100 & 1302 & 4818 & 0.44 & 0.45  & 0.45 & 0.45  \\ \hline
150 & 1952 & 7218 & 0.61 & 0.63  & 0.62 & 0.65  \\ \hline
200 & 2602 & 9618 & 0.84 & 0.85  & 0.85 & 0.84  \\ \hline
300 & 3902 & 14418 & 1.23 & 1.25  & 1.25 & 1.25  \\ \bottomrule
\end{tabular}
\end{adjustbox}
\caption{Sanity check on the BitVec case study - 1st Architecture}
\label{tab:Table 2}
\end{table}

\withLtwo{
\begin{table}[h]
\begin{adjustbox}{center}
\begin{tabular}{cccccccc}
\toprule
\multicolumn{1}{p{2cm}}{\centering \textbf{\#nodes \\per  layer}} &
\multicolumn{1}{p{2cm}}{\centering \textbf{\# \\params}} &
\multicolumn{1}{p{2cm}}{\centering \textbf{\# SMT \\ variables}} &   
\multicolumn{1}{p{2cm}}{\centering \textbf{Strict \\ Equiv.}} & 
\multicolumn{1}{p{2cm}}{\centering\textbf{ $L_{1}$} \\ 
\textbf{Equiv.}} & \multicolumn{1}{p{2cm}}{\centering \textbf{$L_{2}$} \\ 
\textbf{Equiv.}} & \multicolumn{1}{p{2cm}}{\centering \textbf{$L_{\infty}$} \\ 
\textbf{Equiv.}} & \multicolumn{1}{p{2cm}}{\centering \textbf{Argmax} \\ \textbf{Equiv.}} \\ \hline
5 & 97 & 378 & 0.04 & 0.04 & 0.04 & 0.04 & 0.04  \\ \hline
10 & 242 & 938 & 0.1 & 0.09 & 0.09 & 0.09 & 0.1  \\ \hline 
15 & 437 & 1698 & 0.15 & 0.15 & 0.16 & 0.16 & 0.15 \\ \hline
20 & 682 & 2658 & 0.24 & 0.24 & 0.24 & 0.23 & 0.23  \\ \hline
30 & 1322 & 5178 & 0.4 & 0.39 & 0.4 & 0.39 & 0.42  \\ \hline
40 & 2162 & 8498 & 0.62 & 0.62 & 0.63 & 0.63 & 0.63  \\ \hline
50 & 3202 & 12618 & 0.87 & 0.91 & 0.89 & 0.88 & 0.92  \\ \hline
60 & 4442 & 17538 & 1.17 & 1.2 & 1.17 & 1.16 & 1.23  \\ \bottomrule
\end{tabular}
\end{adjustbox}
\caption{Sanity check on the BitVec case study - 2nd Architecture}
\label{tab:Table 3}
\end{table}}

\begin{table}[h]
\resizebox{\textwidth}{!}{
\begin{tabular}{cccccccc}
\toprule
\multicolumn{1}{p{2cm}}{\centering \textbf{\#nodes \\per  layer}} &
\multicolumn{1}{p{2cm}}{\centering \textbf{\# \\params}} &
\multicolumn{1}{p{2cm}}{\centering \textbf{\# SMT \\ variables}} &   
\multicolumn{1}{p{2cm}}{\centering \textbf{Strict \\ Equiv.}} & 
\multicolumn{1}{p{2cm}}{\centering\textbf{ $L_{1}$} \\ 
\textbf{Equiv.}} & \multicolumn{1}{p{2cm}}{\centering \textbf{$L_{\infty}$} \\ 
\textbf{Equiv.}} & \multicolumn{1}{p{2cm}}{\centering \textbf{Argmax} \\ \textbf{Equiv.}} \\ \hline
5 & 97 & 378 & 0.04 & 0.04 & 0.04  & 0.04  \\ \hline
10 & 242 & 938 & 0.1 & 0.09 & 0.09  & 0.1  \\ \hline 
15 & 437 & 1698 & 0.15 & 0.15 & 0.16 & 0.15 \\ \hline
20 & 682 & 2658 & 0.24 & 0.24 & 0.23 & 0.23  \\ \hline
30 & 1322 & 5178 & 0.4 & 0.39 & 0.39 & 0.42  \\ \hline
40 & 2162 & 8498 & 0.62 & 0.62 & 0.63 & 0.63  \\ \hline
50 & 3202 & 12618 & 0.87 & 0.91  & 0.88 & 0.92  \\ \hline
60 & 4442 & 17538 & 1.17 & 1.2  & 1.16 & 1.23  \\ \bottomrule
\end{tabular}
}
\caption{Sanity check on the BitVec case study - 2nd Architecture}
\label{tab:Table 3}
\end{table}

\begin{table}[h!]
\resizebox{\textwidth}{!}{
\begin{tabular}{cccccccc}
\toprule
\multicolumn{1}{p{2cm}}{\centering \# \textbf{nodes \\per  layer}} &
\multicolumn{1}{p{2cm}}{\centering \textbf{\# \\params}} &
\multicolumn{1}{p{2cm}}{\centering \textbf{\# SMT \\ variables}} &   
\multicolumn{1}{p{2cm}}{\centering \textbf{Strict \\ Equiv.}} & 
\multicolumn{1}{p{2cm}}{\centering $L_{1}$ \\ 
\textbf{Equiv.}} & \multicolumn{1}{p{2cm}}{\centering {$L_{\infty}$} \\ 
\textbf{Equiv.}} & \multicolumn{1}{p{2cm}}{\centering\textbf{ Argmax \\ Equiv.}} \\ \hline
10 & 7960 & 32424 & 2.5 & 2.53 & 2.63  & 2.67  \\ \hline 
30 & 23860 & 95624 & 7.52 & 7.73 &  7.63 & 7.5 \\ \hline
50 & 39760 & 158824 & 12.2 & 12.4  & 12.8 & 12.4  \\ \hline
100 & 79510 & 316824 & 24.5 & 24.3  & 25 & 24.6  \\ \hline
200 & 159010 & 632824 & 48.4 & 55.1  & 48.8 & 48.2  \\ \hline
300 & 238510 & 948824 & 74 & 74 & 75 & 73  \\ \hline
500 & 397510 & 1580824 & 121 & 124  & 128 & 119  \\ \hline
750 & 596260 & 2370824 & 182 & 193  & 203 & 182  \\ \hline
1000 & 795010 & 3160824 & 241 & 247  & 257 & 256  \\ \hline
1300 & 1033510 & 4108824 & 314 & 336  & 331 & 321  \\ \hline
1700 & 1351510 & 5372824 & 420 & 434  & 435 & 437  \\ \hline
2000 & 1590010 & 6320824 & 467 & 512  & 492 & 508  \\ \bottomrule
\end{tabular}
}
\caption{Sanity check on the MNIST case study - 1st Architecture}
\label{tab:Table 4}
\end{table}

\begin{table}[h]
\resizebox{\textwidth}{!}{
\begin{tabular}{cccccccc}
\toprule
\multicolumn{1}{p{2cm}}{\centering \textbf{\# nodes \\per layer}} &
\multicolumn{1}{p{2cm}}{\centering \textbf{\# params}} &
\multicolumn{1}{p{2cm}}{\centering \textbf{\# SMT \\ variables}} &   
\multicolumn{1}{p{2cm}}{\centering \textbf{Strict \\ Equiv.}} & 
\multicolumn{1}{p{2cm}}{\centering \textbf{$L_{1}$ \\ 
Equiv.}} & \multicolumn{1}{p{2cm}}{\centering \textbf{$L_{\infty}$ \\ 
Equiv.}} & \multicolumn{1}{p{2cm}}{\centering \textbf{Argmax \\ Equiv.}} \\ \hline
10 & 8070 & 32864 & 2.7 & 2.7  & 2.54 & 2.57  \\ \hline 
30 & 24790 & 99344 & 7.55 & 7.86  & 7.5 & 7.5 \\ \hline
50 & 42310 & 169024 & 13.2 & 12.9  & 13.7 & 12.9  \\ \hline
100 & 89610 & 357224 & 26.5 & 27.5 &  26 & 27.7  \\ \hline
200 & 199210 & 793624 & 58.7 & 62.2 &  61.6 & 64.4  \\ \hline
300 & 328810 & 1310024 & 99 & 100 & 99  & 101  \\ \hline
500 & 648010 & 2582824 & 194 & 194  & 192 & 192  \\ \hline
750 & 1159510 & 4623824 & 334 & 340  & 354 & 349  \\ \hline
1000 & 1796010 & 7164824 & 524 & 530  & 523 & 560  \\ \hline
1300 & 2724810 & 10874024 & 797 & 779  & 836 & 857  \\ \hline
1700 & 4243210 & 16939624 & 1225 & 1161  & 1237 & 1223 \\ \hline
2000 & 5592010 & 22328824 & 1435 & 1530  & 1549 & 1581  \\ \bottomrule
\end{tabular}
}
\caption{Sanity check on the MNIST case study - 2nd Architecture}
\label{tab:Table 5}
\end{table}

In all these sanity checks the SMT solver returned UNSAT, which correctly indicates that the formula is unsatisfiable, i.e., that the two (identical) neural networks are equivalent, as expected.

\subsection{Equivalence checking case studies}
\label{exp2}

In our second set of experiments, we conduct equivalence checking on neural networks with different architectures. On the one hand, we aim to test the efficiency of our proposed methodology regarding the equivalence relations that can be checked for the models we have trained. On the other hand, we want to identify the computation time limitations and connect them with the total number of SMT variables that are being verified per equivalence check. For that reason we set a time limit of $10$ minutes on the solver to respond whether the assignment (query) is SAT (satisfiable) or UNSAT (unsatisfiable). Recall that we have structured the encoding in such a way so that we expect the solver to return UNSAT in case the equivalence holds, and SAT otherwise.  

\subsubsection{Experiments with NN classifiers}

In this first subsection we present the results regarding the equivalence checking of Neural Networks that serve as classifiers, so we include material relevant to the BitVec and MNIST case studies. For the former we have checked one by one all the neural network pairs between the two different architectures (e.g. $1^{\text{st}}$ model of Table~\ref{tab:Table 2} versus $1^{\text{st}}$ model of Table~\ref{tab:Table 3}), under all equivalence relation criteria. The aforementioned experimental results are depicted in Table~\ref{tab:Table 6}. The first column shows the neural network pair that is checked while the second the total number of SMT variables to be checked for this particular pair. In the next columns we present the answer that the solver returned and the time it took (in seconds) to respond. For the $\epsilon$-approximate equivalence relations we note the value of $\epsilon$ under which the neural networks checked. We choose big values of $\epsilon$ because our models, do not contain any non-linear activation function on the output layer (e.g. Softmax, Sigmoid etc.), so there is no guarantee that the outputs would be scaled on the same value ranges. Apparently, there is an extra term in the tables namely \textbf{MME}, that stands for Maximum Memory Exceeded and is the reason why the solver fails to return an answer before reaches the time limit of $10$ minutes.

\vspace{3mm}

\begin{table}[h]
\resizebox{\textwidth}{!}{
\begin{tabular}{@{}cccccccc@{}}
\toprule
\multicolumn{1}{p{2cm}}{\centering\textbf{ Model Pairs}}  &
\multicolumn{3}{c}{\centering \textbf{\# SMT variables} } &
\multicolumn{1}{p{2cm}}{\centering\textbf{ Strict \\ Equiv.}} & 
\multicolumn{1}{p{2cm}}{\centering\textbf{ $L_{1} > 5$}} &
\multicolumn{1}{p{2cm}}{\centering\textbf{ $L_{\infty} > 10$}} & \multicolumn{1}{p{2cm}}{\centering \textbf{Argmax \\ Equiv.}} \\ 
 &\textbf{\textit{ Input}} & \textbf{\textit{Internal}} & \textbf{\textit{Output}} & & &  & \\
\hline
model$\_1\_1$ vs model$\_2\_1$ & 10 & 424 & 4 & SAT/$0.042$ s & UNSAT/$35$ s  & UNSAT/$48$ s & SAT/$0.23$  s \\ \hline
model$\_1\_2$ vs model$\_2\_2$ & 10 & 944 & 4 & SAT/$0.075$ s & SAT/$0.20$ s  & UNSAT/$157$ s & SAT/$0.4$ s \\ \hline
model$\_1\_3$ vs model$\_2\_3$ & 10 & 1630 & 4 & SAT/$0.124$ s & UNSAT/$385$ s   & UNSAT/$531$ s & UNSAT/$182$ s  \\ \hline
model$\_1\_4$vs model$\_2\_4$ & 10 & 2524 & 4 & SAT/$0.19$ s & SAT/$245$ s  & Timeout& SAT/$86$ s \\ \hline
model$\_1\_5$ vs model$\_2\_5$ & 10 & 4984 & 4 & SAT/$0.35$ s & Timeout  & MME/$509$ s & SAT/$240$ s \\ \hline
model$\_1\_6$ vs model$\_2\_6$ & 10 & 7844 & 4 & SAT/$0.5$ s & Timeout  & MME/$568$ s & SAT/$450$ s \\ \hline
model$\_1\_7$ vs model$\_2\_7$ & 10 & 11104 & 4 & SAT/$0.75$ s & Timeout  & MME/$588$ s & Timeout  \\ \hline
model$\_1\_8$ vs model$\_2\_8$ & 10 & 15964 & 4 & SAT/$1.13$ s & Timeout  & Timeout & Timeout  \\ \bottomrule
\end{tabular}
}
\caption{Equivalence checking on the BitVec case study}
\label{tab:Table 6}
\end{table}

Now, in Table~\ref{tab:Table 7} the corresponding results for the MNIST case study are presented. The Table structure is identical with Table~\ref{tab:Table 6}, so there is no need for extra comments regarding the interpretability of the results. Here, in order to avoid states explosion due to the very big number of SMT variables that are being used, we have included the first five pairs of MNIST models as shown in Table~\ref{tab:Table 4} and Table~\ref{tab:Table 5}. 

\withLtwo{
\begin{table}[h]
\resizebox{\textwidth}{!}{
\begin{tabular}{ccccccccc}
\toprule
\multicolumn{1}{p{2cm}}{\centering\textbf{ Model Pairs}}  &
\multicolumn{3}{c}{\centering \textbf{\# SMT variables} } &
\multicolumn{1}{p{2cm}}{\centering \textbf{Strict \\ Equiv.}} & 
\multicolumn{1}{p{2cm}}{\centering \textbf{$L_{1} > 5$}} &
\multicolumn{1}{p{2cm}}{\centering\textbf{ $L_{2} > 25$}} &
\multicolumn{1}{p{2cm}}{\centering \textbf{$L_{\infty} > 10$}} & \multicolumn{1}{p{2cm}}{\centering \textbf{Argmax \\ Equiv.}} \\ 
 & \textbf{\textit{Input}} & \textbf{\textit{Internal}} & \textbf{\textit{Output}} & & & & & \\
\hline
mnist$\_1\_1$ vs mnist$\_2\_1$ & 784 & 31840 & 20 & SAT/$2.1$ s & SAT/$44$ s & Timeout & SAT/$42$ s & SAT/$41$ s  \\ \hline
mnist$\_1\_2$ vs mnist$\_2\_2$ & 784 & 96680 & 20 & SAT/$6.2$ s & SAT/$16$ s & Timeout & SAT/$17$ s & SAT/$17$ s \\ \hline
mnist$\_1\_3$ vs mnist$\_2\_3$ & 784 & 163120 & 20 & SAT/$10.3$ s & SAT/$28$ s & MME/$575$ s & SAT/$28$ s & MME/$230$ s \\ \hline
mnist$\_1\_4$ vs mnist$\_2\_4$ & 784 & 336220 & 20 & SAT/$21$ s & SAT/$56$ s & MME/$95$ s & SAT/ $57$ s & SAT/$54$  s \\ \hline
mnist$\_1\_5$ vs mnist$\_2\_5$ & 784 & 712420 & 20 & SAT/$45$ s & SAT/$120$ s & MME/$154$ s & SAT/$120$ s & SAT/$118$ s \\ \bottomrule
\end{tabular}
}
\caption{Equivalence checking on MNIST case study}
\label{tab:Table 7}
\end{table}}
\begin{table}[h]
\resizebox{\textwidth}{!}{
\begin{tabular}{cccccccc}
\toprule
\multicolumn{1}{p{2cm}}{\centering\textbf{ Model Pairs}}  &
\multicolumn{3}{c}{\centering \textbf{\# SMT variables} } &
\multicolumn{1}{p{2cm}}{\centering \textbf{Strict \\ Equiv.}} & 
\multicolumn{1}{p{2cm}}{\centering \textbf{$L_{1} > 5$}} &
\multicolumn{1}{p{2cm}}{\centering \textbf{$L_{\infty} > 10$}} & \multicolumn{1}{p{2cm}}{\centering \textbf{Argmax \\ Equiv.}} \\ 
 & \textbf{\textit{Input}} & \textbf{\textit{Internal}} & \textbf{\textit{Output}} & & & & \\
\hline
mnist$\_1\_1$ vs mnist$\_2\_1$ & 784 & 31840 & 20 & SAT/$2.1$ s & SAT/$44$ s &  SAT/$42$ s & SAT/$41$ s  \\ \hline
mnist$\_1\_2$ vs mnist$\_2\_2$ & 784 & 96680 & 20 & SAT/$6.2$ s & SAT/$16$ s &  SAT/$17$ s & SAT/$17$ s \\ \hline
mnist$\_1\_3$ vs mnist$\_2\_3$ & 784 & 163120 & 20 & SAT/$10.3$ s & SAT/$28$ s  & SAT/$28$ s & MME/$230$ s \\ \hline
mnist$\_1\_4$ vs mnist$\_2\_4$ & 784 & 336220 & 20 & SAT/$21$ s & SAT/$56$ s & SAT/ $57$ s & SAT/$54$  s \\ \hline
mnist$\_1\_5$ vs mnist$\_2\_5$ & 784 & 712420 & 20 & SAT/$45$ s & SAT/$120$ s &  SAT/$120$ s & SAT/$118$ s \\ \bottomrule
\end{tabular}
}
\caption{Equivalence checking on the MNIST case study}
\label{tab:Table 7}
\end{table}

\subsubsection{Experiments with regression models}

In this subsection we present the results of the regression problem, particularly the equivalence checking of neural networks that serve as MPC controllers for lane keeping assistant systems. More details for the case study that we have reproduced can be found in \footnote{https://www.mathworks.com/help/reinforcement-learning/ug/imitate-mpc-controller-for-lane-keeping-assist.html}. Here, we alternate our experimental setup compared to the one we used in classification problems. More specifically, instead of verifying two different neural network architectures, we check the same neural network in different versions. The versions is an outcome of the number of epochs that the model is trained before being verified. We experiment with three versions of the MPC controller trained for $30, 35$ and $40$ epochs. Since, there is only one output variable on the specific problem, we do not include the Argmax equivalence relation on the experiments. Additionally, for the aforementioned reason, we present results only for the $L_{1}$ norm because there is not any distinction between different $L_{p}$ norms if they are implemented on scalar values. Table~\ref{tab:Table 8} concentrates the relevant results.

\begin{table}[h]
\begin{adjustbox}{center}
\begin{tabular}{@{}cccccc@{}}
\toprule
\multicolumn{1}{p{2cm}}{\centering\textbf{ Model Pairs}} & 
\multicolumn{3}{c}{\centering \textbf{\# SMT variables}} & 
\multicolumn{1}{p{2.5cm}}{\centering \textbf{Strict Equivalence}} & 
\multicolumn{1}{p{1.5cm}}{\centering \textbf{$L_{1}>0.5$}} \\
 & \textit{\textbf{Input}} & \textbf{\textit{Internal}} & \textit{\textbf{Output}} & & \\
\hline
MPC$\_30$ vs MPC$\_35$ & 6 & 17912 & 2 & SAT/$1.95$ s & Timeout \\ \hline
MPC$\_30$ vs MPC$\_40$ & 6 & 17912 & 2 & SAT/$1.97$ s & Timeout \\ \hline
MPC$\_35$ vs MPC$\_40$ & 6 & 17912 & 2 & SAT/$1.98$ s & Timeout \\ \bottomrule
\end{tabular}
\end{adjustbox}
\caption{Equivalence checking on the Regression case study}
\label{tab:Table 8}
\end{table}

\subsubsection{Experiments with weight perturbations}

One remarkable finding derived from the experiments in section~\ref{exp2} is that it takes much less time for the solver to find a SAT assignment compared to an UNSAT one. This is partially justified by the fact that it is much simpler (for the solver) to find a counterexample (SAT case) opposed to exhaustively search for all possible input combinations that do not violate an assignment (UNSAT case). In this last part of the experiments, our primary goal is to "push" the SMT solver in order to respond with UNSAT. With this goal in mind, we first choose one of the trained models on MNIST and then we randomly alter the values of some parameters and save a copy of that neural network. Then we deploy equivalence checking between these two model versions. As shown in Table~\ref{tab:Table 9}, as soon as we alter two parameters (out of $7960$) the solver reaches the $10$ min Timeout set.  

\begin{table}[h]
\resizebox{\textwidth}{!}{
\begin{tabular}{ccccccccc}
\toprule
\multicolumn{1}{p{2cm}}{\centering \textbf{Model}}  &
\multicolumn{3}{c}{\centering \textbf{\# SMT variables} } &
\multicolumn{1}{p{2cm}}{\centering \textbf{\# weight changes}} & 
\multicolumn{1}{p{2cm}}{\centering\textbf{ Value range}} &
\multicolumn{1}{p{2cm}}{\centering \textbf{$L_{1} > 5$}} &
\multicolumn{1}{p{2cm}}{\centering \textbf{$L_{\infty} > 10$}} & 
\multicolumn{1}{p{2cm}}{\centering\textbf{ Argmax \\ Equiv.}} \\ 
 & \textbf{\textit{Input}} & \textbf{\textit{Internal}} & \textit{\textbf{Output}} & & & & & \\
\hline
mnist$\_1\_1$ & 784 & 31840 & 20 & $1$ & $1\mathrm{e}{-1}$ - $1\mathrm{e}{-6}$ & UNSAT/$30$ s & UNSAT/$30$ s & Timeout  \\ \hline
mnist$\_1\_1$ & 784 & 31840 & 20 & $2$ & $1\mathrm{e}{-1}$ - $1\mathrm{e}{-6}$ & Timeout & Timeout & Timeout \\ \bottomrule
\end{tabular}
}
\caption{Equivalence checking with weight perturbations on the MNIST case study}
\label{tab:Table 9}
\end{table}

\section{Related work}
\label{sec:6}

Verification of neural networks with respect to various correctness properties (e.g. safety/reachability, robustness etc.) is an area of fast growing interest due to the many interesting and often critical applications, in which neural networks are employed. Two comprehensive surveys of all these related works, including the verification of neural networks using SMT solvers, are given in~\cite{leofante2018automated} and~\cite{HUANG2020100270}.

Regarding the equivalence verification of two neural networks, \cite{narodytska2018verifying} is probably the only related work that focuses on this problem. This work is also based on a SAT/SMT based encoding of the equivalence checking problem, but the overall approach is applicable only to a specific category of neural networks, the so-called binarized neural networks~\cite{courbariaux2016binarized} that are not widely used in many different real-life applications. Worth to mention is the work in \cite{Paulsen_2020}, where the authors focus on analyzing the relationship between two neural networks, e.g. whether a modified version of an existing neural network produces outputs within some bound relative to the original network. While the focus is not to answer the question of equivalence given an appropriate equivalence criterion, the authors propose an interesting ``differential verification'' technique that consists of a forward interval analysis through the network's layers, followed by a backward pass that iteratively refines the approximation, until having verified the property of interest.

In the quest of more clever and potentially scalable encodings for our problem, we are going to study \cite{singh2019abstract}, where the authors propose a methodology based on abstractions of the input domain using Zonotopes and Polyhedra, along with using an MILP solver for verifying properties of neural networks of varying complexity and size. Another interesting source is \cite{10.1007/978-3-319-63387-9_5}, which presents an SMT-based verification method for the verification of a single neural network. However, the applicability of that work is limited only to ReLU-based neural networks. Finally, an interesting symbolic representation targeting only piecewise linear neural networks is the one presented in~\cite{Sotoudeh19}.



\section{Conclusions}

In this work, we formally defined the equivalence checking problem for neural networks and we introduced a series of equivalence criteria that might be more appropriate than others for specific applications and verification requirements. Furthermore, we provided a first SMT-based encoding of the equivalence checking problem, as well as experimental results that demonstrate its sanity and give insight into its current scalability limitations.  

In our future research plans, we aim to explore whether the equivalence checking problem (and our equivalence criteria) can be encoded in state-of-the-art verification tools for neural networks (e.g. Reluplex~\cite{10.1007/978-3-319-63387-9_5}, ERAN~\cite{singh2019abstract}, $\alpha - \beta$ Crown~\cite{10.5555/3327345.3327402, xu2021fast, wang2021beta}, VNN competition~\cite{bak2021second}) through the parallel composition of the two networks that are to be compared.

As additional research priorities, we also intend to explore the scalability margins of alternative solution encodings, including an optimized version of our current encoding (by elimination of the internal SMT variables) and a mixed-integer linear programming encoding. Lastly, it may be also worth to explore the practical effectiveness of technical solutions to similar problems from other fields, like for example the equivalence checking of digital circuits~\cite{915010, 10.1145/1233501.1233679, 4196156}. In this context, we may need to rely on novel ideas towards the layer-by-layer checking of equivalence between two neural networks.

\bibliographystyle{plain}
\bibliography{NNeqBibl}

\end{document}